\documentclass[runningheads]{llncs}
\usepackage{graphicx}
\usepackage{amsmath,amssymb} 
\usepackage{color}
\newcommand{\etal}{\textit{et al}. }
\newcommand{\ie}{\textit{i}.\textit{e}., }
\newcommand{\eg}{\textit{e}.\textit{g}., }
\usepackage{amssymb}
\usepackage{pifont}

\usepackage{subcaption}     
\usepackage[utf8]{inputenc} 
\usepackage[T1]{fontenc}    
\usepackage{hyperref}       
\usepackage{url}            
\usepackage{booktabs}       
\usepackage{amsfonts}       
\usepackage{nicefrac}       
\usepackage{microtype}      
\usepackage{epsfig}
\usepackage{graphicx}
\hypersetup{colorlinks=true, linkcolor=red}
\usepackage{wrapfig}
\usepackage{algorithm}
\usepackage{algorithmic}

\usepackage[sort&compress,square,comma,numbers]{natbib}
\makeatletter
\renewcommand\bibsection%
{
  \section*{\refname
    \@mkboth{\MakeUppercase{\refname}}{\MakeUppercase{\refname}}}
}
\makeatother
\usepackage{array}
\newcolumntype{L}[1]{>{\raggedright\let\newline\\\arraybackslash\hspace{0pt}}m{#1}}
\newcolumntype{C}[1]{>{\centering\let\newline\\\arraybackslash\hspace{0pt}}m{#1}}
\newcolumntype{R}[1]{>{\raggedleft\let\newline\\\arraybackslash\hspace{0pt}}m{#1}}

\begin{document}
\pagestyle{headings}
\mainmatter

\def\ACCV18SubNumber{392}  

\title{Deep-HR: Fast Heart Rate Estimation from Face Video Under Realistic Conditions} 
\titlerunning{Deep-HR: Fast Heart Rate Estimation from Face Video Under Realistic Conditions}
\authorrunning{Sabokrou et al.}

\author{Mohammad Sabokrou$^1$, Masoud Pourreza, Xiaobai Li$^2$ \\ Mahmood Fathy$^{3}$, Guoying Zhao$^{2}$}
\institute{$^1$Institute for Research in Fundamental Sciences (IPM)\quad$^2$University of Oulu\quad$^3$Iran University of Science and Technology}
\maketitle

\begin{abstract}
This paper presents a novel method for remote heart rate (HR)
estimation. Recent studies have proved that blood pumping
by the heart is highly correlated to the intense color of face pixels, and surprisingly
can be utilized for remote HR estimation.  Researchers successfully proposed several methods for this task, but making it work in realistic situations is still a challenging problem in computer vision community. Furthermore, learning to solve such a complex task on a dataset with very limited annotated samples is not reasonable. Consequently, researchers do not prefer to use the deep learning approaches for this problem. In this paper,  we propose a simple yet efficient approach to benefit the advantages of the Deep Neural Network (DNN) by simplifying  HR estimation from a complex task to learning from very correlated representation to HR. Inspired by  previous work, we learn a component called Front-End (\textbf{FE}) to provide a discriminative representation of face videos, afterward a light deep regression auto-encoder as Back-End (\textbf{BE}) is learned to map the \textbf{FE} representation to HR. Regression task on the informative representation is simple and could be learned efficiently on limited training samples. Beside of this, to be more accurate and work well on low-quality videos,  two deep encoder-decoder networks are trained to refine the output of \textbf{FE}.  We also introduce a challenging dataset (HR-D) to show that our method can efficiently work in realistic conditions. Experimental results on HR-D
and MAHNOB datasets confirm that our method could run as a real-time method and estimate the average HR better than state-of-the-art ones.  
\end{abstract}

\section{Introduction}

Remote HR estimation  from face video is an interesting and challenging task. HR is closely related  to the important tasks and  could be used in a wide range of applications such  as health-care\cite{hung2004wearable},  face spoofing detection~\cite{li2016generalized,hernandez2018time}, fitness assessment~\cite{vspetlikvisual}, and  emotion recognition \cite{li2014remote}.

\begin{figure}[t]
    \centering
   \includegraphics[width=0.7\linewidth]{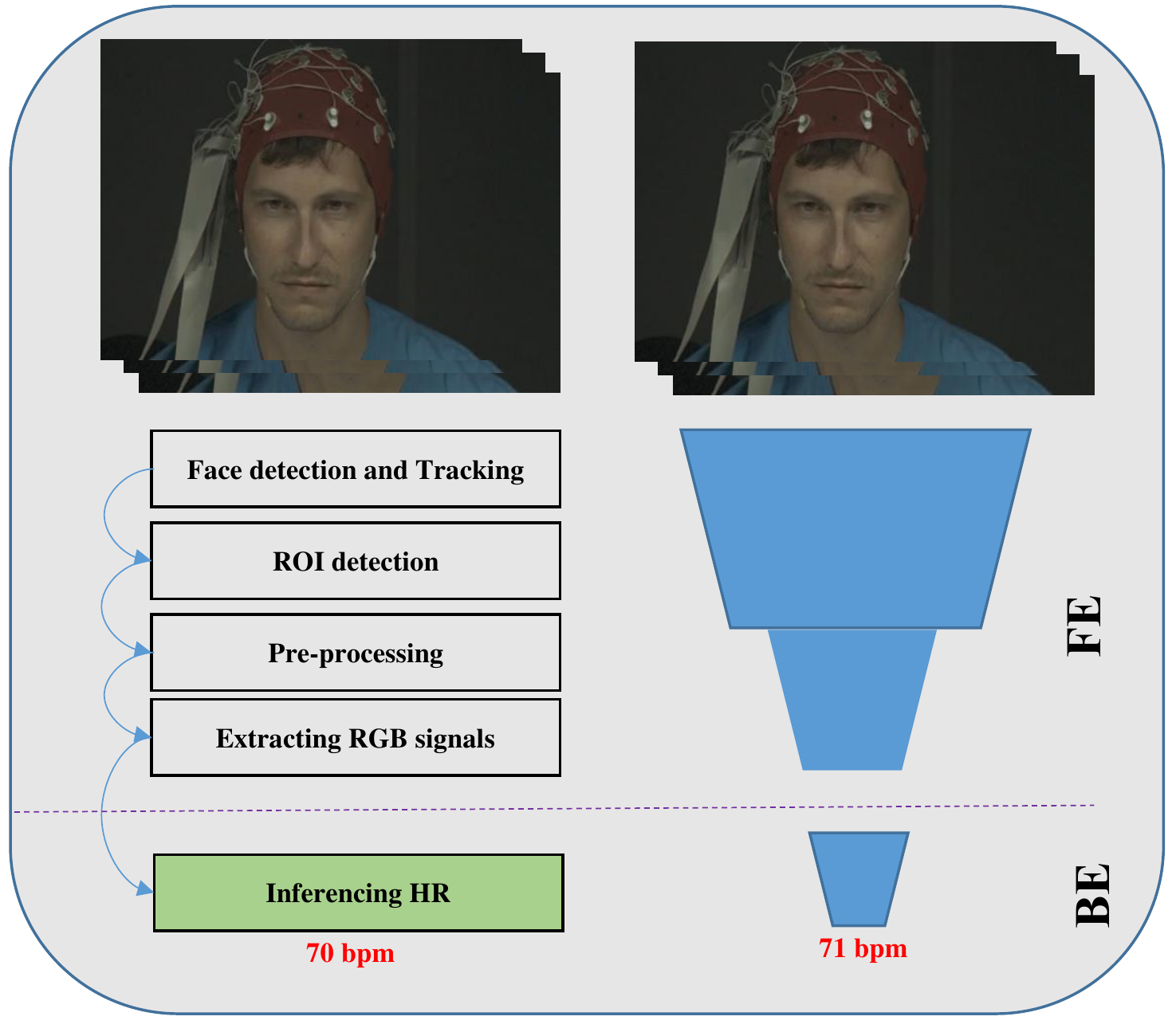}
    \caption{Deep network vs traditional paradigm for HR rate estimation. {\it Left:} An Example of traditional method. These methods contain  several sub-tasks such as face detection, face tracking, and RoI detection.  {\it Right:} Our proposed frame-work based on deep learning method.  }
    \label{fig:deepvstraditional}
\end{figure}
Recently, this topic gains more attention and several effective methods have been proposed~\cite{li2014remote,tulyakov2016self,lam2015robust}.  None of the previous proposed state-of-the-art methods (Except \cite{niu2018synrhythm,niu2017continuous,vspetlikvisual}) for remote HR estimation are devised based  on data-driven method.  Face video representation using hand-crafted features, and applying signal processing techniques on such representation are two main components of the previous methods whereas these components are designed based on some assumptions and restrictions. Traditional methods have achieved very good performance on videos captured in a controlled environment while they are not robust against noise, face movement, and illumination changes.   Besides, due to  the nature of  applications that are based on   HR estimation, real time
processing is necessary, while most of the previous methods for remote  HR estimation are too complex to run in real time.  

Recently, data-driven methods especially those which are based on deep learning, have achieved the promising  results to learn complex tasks such as video anomaly detection \cite{sabokrou2018adversarially,sabokrou2018avid} and image classification~\cite{he2016deep}. Deep learning methods, \eg,  convolutional Neural networks (CNNs), need a lot of training samples to work well, while there is not any  public dataset for HR estimation task with plenty of  annotated videos. Consequently, proposing a deep learning approach for estimating the HR is not straightforward.

Based on the conventional methods \cite{li2014remote} for HR estimation, a pre-defined subsequent  processes such as RoI ({\it region of interest}) detection should be done on videos. On the other side,  different to conventional methods, for machine learning algorithms big number of training samples for  building  an efficient  learning model is very important.  As mentioned,  HR estimation from facial videos is composed of different processes such as  face detection, face tracking, RoI  detection,  and informative signal extraction from RoI.  Due to complexity of these tasks, great effort on a lot of annotated data is needed for learning.     Briefly, it can be said that (1) limited number of labeled samples, (2) poor performance on realistic conditions and (3)high complexity of the task, are three main challenges in this filed.   Interesting questions arising out of this problem are:  How could we learn a  DNN  when the labeled training samples are not enough? How could we infer HR very fast? 

Regarding previous studies on remote HR estimation, there are valuable prior-knowledge which can be used  for training a deep structure. On the other hand, the first layers of a deep network are very related to input space and the end layers are influenced by the task (output). According to the points mentioned,  we propose to use only the labeled video clips for the latter steps (last layers) of the methods.     Fig.~\ref{fig:deepvstraditional} shows the difference between our method and previous methods. Contrary to conventional remote HR estimation,  this paper proposes a DNN method for detecting (or generating) the RoI of the face, and extracting the informative features related to HR task. RoI detection and representation  can be considered as two general purpose methods which are independent to  HR estimation task. Consequently, it does not need to train on labeled HR samples and  is called Front-End (\textbf{FE}) of our proposed structure. Thereafter a  supervised fully-connected network is exploited for estimating the  HR from an output of \textbf{FE}. This fully-connected network plays as the Back-End (\textbf{BE}) of our deep structure and  is learned on labeled training samples.

It is worth mentioning that,  HR  is highly dependant on the subtle color changes of face skin. Generally, in real-world applications, the recorded videos are contaminated by noise which brings additional  difficulty to track the subtle color changes in face.  Inspired by the recent developments in generative adversarial networks (GANs) \cite{goodfellow2014generative} and  for being robust against noise and enhancing the quality of  the videos frames, two encoder-decoder networks $\mathcal{G}_1$ and $\mathcal{G}_2$ are adversarially learned  for refining the  output of  \textbf{FE}.

In summary, the main contributions of this paper are as follows: (1) We  propose a  simple yet effective remote HR estimation method based on  deep learning. Our  method is able to learn from limited training samples for this complex task.  To best of our knowledge, this paper is one of the first deep learning based method for HR estimation from facial videos. (2) Our method is robust against noise, face movement and low-quality videos.  Two deep networks are  adverserially learned  to refine (\ie, improve the quality of) the outputs of intermediate process  for HR estimation.   (3) Our results are better than state-of-the-art performance in term of both complexity and accuracy. Deep-HR  efficiently  work in real-time. (4)We  introduce a new challenging dataset named HR-D   and experiments show that  our method is able to perform  promisingly  in a real-world and realistic conditions.  

\begin{figure*}[h]
    \centering
   \includegraphics[width=1\linewidth]{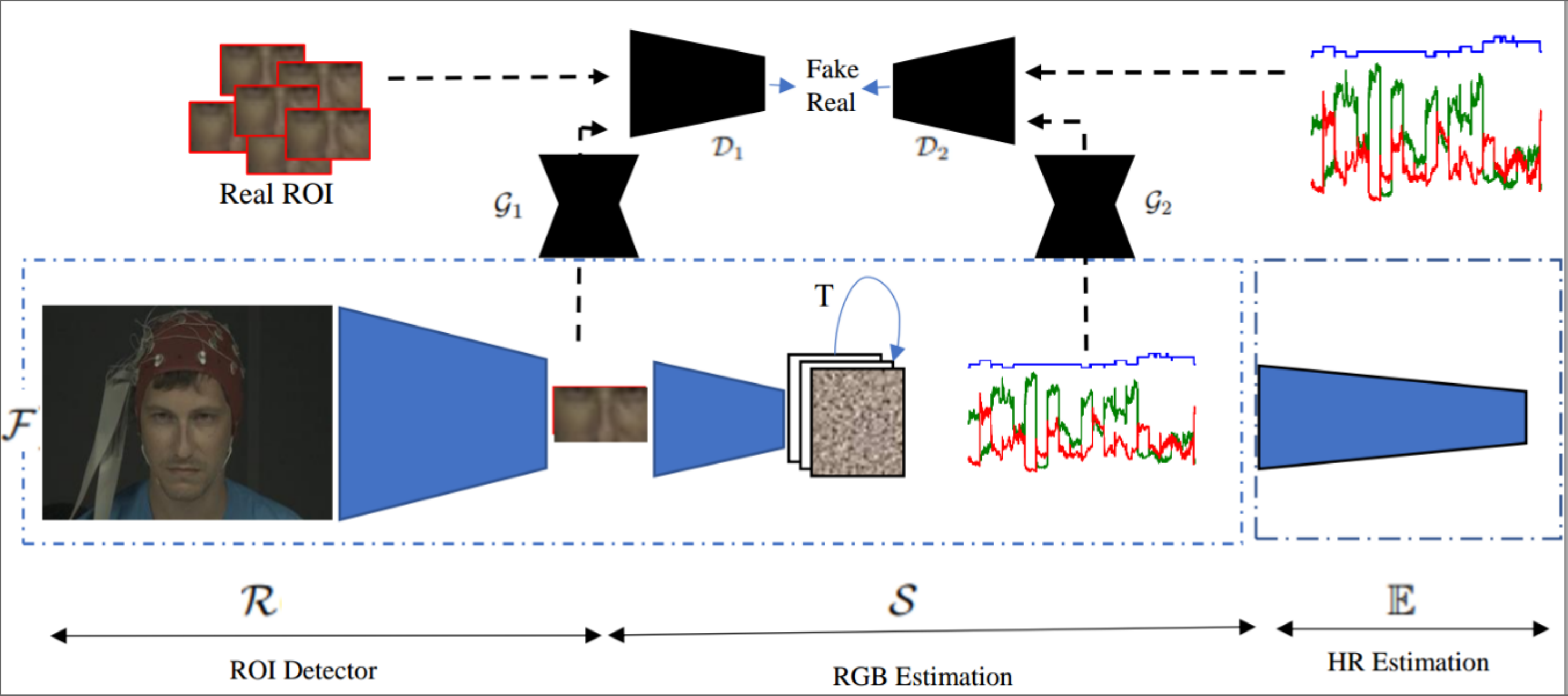}
    \caption{The outline of our method  for HR estimation.  $\mathcal{R}$ detects the RoI of its input~($\mathcal{F}$), $\mathcal{S}$ extracts informative signal from detected RoI by $\mathcal{R}$. $\mathcal{R}(\mathcal{S}(\mathcal{F}))$ is  an informative representation of $\mathcal{F}$.   $\mathbb{E}$ learns on labeled HR samples, to estimate HR from represented face videos (\ie,  $\mathcal{R}(\mathcal{S}(\mathcal{F}))$).  $\mathcal{D}_{1}+\mathcal{G}_{1}$  and $\mathcal{D}_{2}+\mathcal{G}_{2}$ are adversarially learned to enhance  the quality of $\mathcal{R}$ and $\mathcal{S}$ outputs.} 
   \label{fig:os}
\end{figure*}

\section{Related Work}
\label{sec:rw}
Recently remote HR estimation based on  videos is widely attended by  researchers. The earlier methods were focused on investigating a correlation between face videos and heart rate.  
Exploiting face videos for estimating   HR remotely was first investigated by Verkruysse \etal in 2008 \cite{verkruysse2008remote}. Since then, a number of methods have been proposed for remote
HR estimation.

An overview of the state-of-the-art methods for remote heart rate estimation from face video is provided in \cite{hassan2017heart}. 
\cite{de2013robust} proposes a method to estimate and  monitor  the blood volume pulse using a regular
camera. \cite{li2014remote}  proposes a method for heart rate estimation, which  suggests calculating the average of green color of RoIs. They have tracked the face on all frames of video and specified a part of the face as a region of interest. This method essentially  uses tracking to deal with the rigid to deal head motion. This method could not work well in realistic conditions. \cite{lam2015robust} proposes an efficient method based on  skin appearance to cope with the weaknesses in \cite{li2014remote}. A quality metric for pulse rate monitoring based on camera for fitness exercise is developed  in \cite{wang2016quality}. Also in the method presented in \cite{stricker2014non}  the signal of related to R, G and B color channels are extracted on the face area, and independent component analysis (ICA) is then exploited to decompose a multivariate temporal signal into independent
non-Gaussian signals, of which one is expected to be the heart rhythm signal. 
A  chrominance difference representation approach under several assumptions on face motion for remote HR estimation is proposed by \cite{de2013robust}, in which chrominance features based on two orthogonal projections of Red-Green-Blue (RGB) space  reduce the influence of face motion. Kumar  \etal \cite{kumar2015distanceppg}  proposed to extract and fuse the green channel signals
of  different regions of interest using the frequency characteristics. Lam \etal \cite{lam2015robust} have suggested random selection  several patches from the region of interest. They exploited a majority vote rule to decide the final HR estimation. In \cite{tulyakov2016self}  the face is divided into multiple regions of interest, and  with a matrix completion approach  temporal signals are refined. Niu \etal \cite{niu2017continuous}have used a multi-patch region of face for HR estimation and introduced the problem of continuous HR estimation.

 \textbf{HR estimation based on deep  learning:}
 Methods based on deep learning  have achieved promising results in different ranges of computer vision tasks. They could efficiently learn the complicated task by observing many of annotated samples related to such task. As remote HR estimation is a complex task while big number of  annotated training samples are not available for it,  solving this problem based on deep learning  has been rarely touched. 

 Niu \etal  \cite{niu2018synrhythm} have proposed an efficient representation  for HR representation, and  based on this representation they generate many of training samples to overcome the problem of limited training samples.  Even though this work achieves good results, their method is not robust against facial movements and noise. Beside of this, generating many samples is very computationally  expensive and time consuming. 
 In  \cite{vspetlikvisual}    a method based on Convolutional Neural Network for estimation HR rate is presented. They have used two deep networks, one  learned for extracting the informative color signals form RoI, and the other one  mapping the color signals to HR.


\section{Deep-HR: Proposed Method}
\label{sec:pm}
Training a DNN from scratch for learning of a complex task such as remote HR estimation in the absence of enough annotated training samples is not straightforward. Almost the general workflow of all the state-of-the-art methods for remote HR estimation is the same, thus we also comply with the same general workflow, but different from the  previous methods, we propose a machine learning approach for  this task  instead of hendcrafted processing. Our method is composed of  two important components: (1) Front-End:~\textbf{FE}, and (2)Back-End:~\textbf{BE}.  The former  one  includes  two CNNs (1)$\mathcal{R}$, and (2)$\mathcal{S}$. Aiming to learn the informative representation of facial videos, $\textbf{FE}$ is learned independently of HR training samples. The second component, \ie, $\textbf{BE}$   is a low-cost fully-connected neural network for HR estimation ($\mathbb{E}$) that is trained on represented training facial videos from \textbf{FE}.  \textbf{FE} is learned to jointly detect the RoI and represent it.  
 
We have explained earlier that accurate HR estimation  highly depends on the capability of the proposed network on capturing the subtle color changes in the face skin. If the video frames are with poor quality,  the performance  would be degenerated.  As a result,  for adapting to  different  conditions, it is better to guarantee the quality of video frames or quality of the representation of videos which is the input of our \textbf{BE} model. To  this end, inspired by \cite{sabokrou2018adversarially}, we propose two refiner deep networks in an adversarial style to enhance the low-quality imagery.  
  
Fig. \ref{fig:os} shows the outline of the proposed method. The details of \textbf{FE},  \textbf{BE}, and adversarial quality check of the output of $\textbf{FE}$  are explained and discussed in following subsections. 

\subsection{FE structure}
\label{sec:fe}
As mentioned, the \textbf{FE} does not need the labeled HR samples and learns independently of the HR estimation task.  The \textbf{FE}  includes  two deep CNNs: (1)$\mathcal{R}$, and (2) $\mathcal{S}$. These two deep networks are learned to detect the RoI and  extract the highly correlated time-series  color variations of RoI. Furthermore, we refine the output of each of these networks to improve the quality of video representation.   Let $\mathcal{F}$ be our video frames of a face. $\mathcal{S}(\mathcal{R}(\mathcal{F}))$ is a compressed and informative representation of $\mathcal{F}$. Note that $\mathcal{R}(\mathcal{F})$ is the coordinates of RoI, which is then cropped from $\mathcal{F}$. From now on, for simplicity, we assume $\mathcal{R}(\mathcal{F})$ is the RoI, not its coordinates.  Instead of direct estimation of  HR from video face, \textbf{BE} could  be efficiently learned on the output of \textbf{FE}. In other words, the  \textbf{FE} as a prepossessing on videos, can simplify the HR estimation task. 

\begin{figure}[h]
    \centering
   \includegraphics[width=0.7\linewidth]{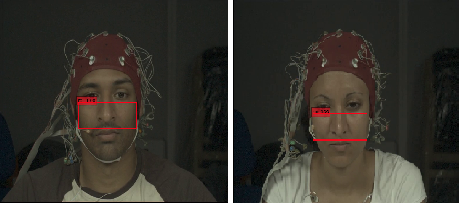}
   \caption{Examples of detected  RoI by $\mathcal{R}$.}
    \label{fig:roi}
\end{figure}

 \begin{figure}[t]
    \centering
   \includegraphics[width=0.6\linewidth]{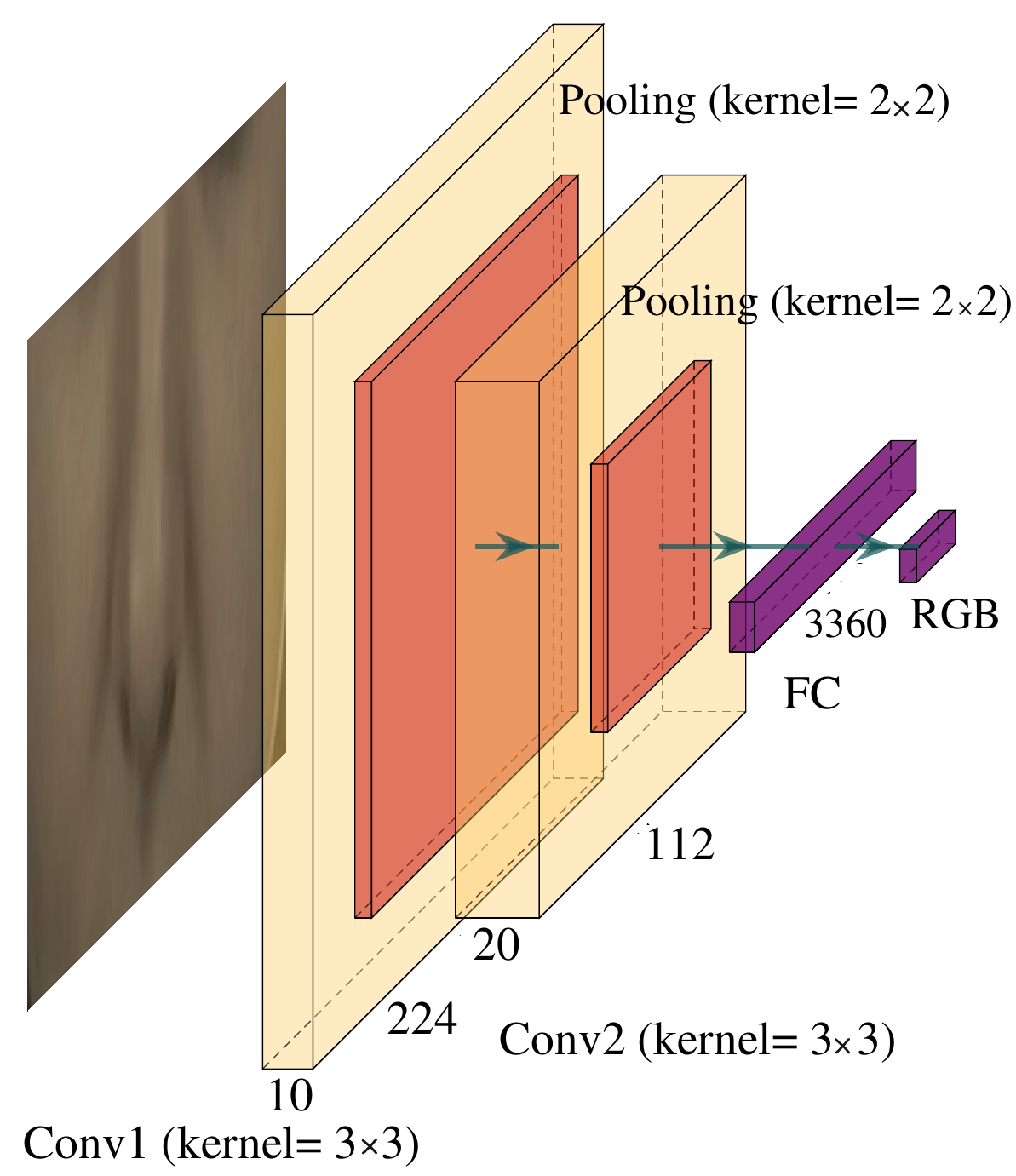}
    \caption{The structure of $\mathcal{S}$ network. This network distills the color information of RoI.  }
    \label{fig:s2}
\end{figure}

\begin{figure*}[t]
    \centering
   \includegraphics[width=1\linewidth]{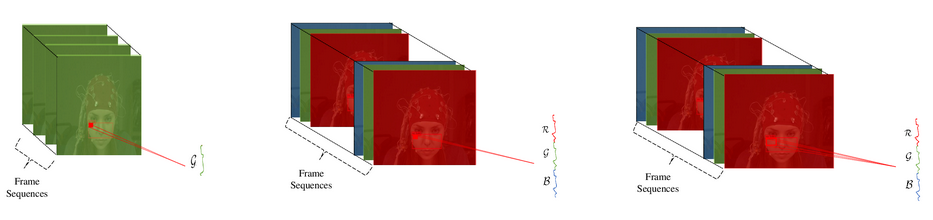}
    \caption{Different strategies for  color signal extraction. {\it Left:} Using G from input image sequences; {\it Middle:} using RGB from input image sequences; {\it Right:} using patch data from input image sequences; }
    \label{fig:rgb_extraction}
\end{figure*}

\textbf{$\mathcal{R}$($\mathcal{R}$oI detector):} 
Researchers have investigated that movements in face such as eye-blinking and facial expressions could have a bad effect on  HR estimation\cite{li2014remote,lam2015robust}. They have found that some regions of the face are more informative and interested in analyzing  HR~\cite{li2014remote}. The overall performance of  HR estimation methods strongly depends on the RoI detection and representation. Ideally,  we would represent the RoI based on features that are robust against facial movements and expressions, while they are also  discriminantive enough to account for the subtle changes in skin color. We propose an efficient and robust CNN named  $\mathcal{R}$ for  RoI detection.

 In respect to the achievement of RFB deep network \cite{liu2017receptive} on accurate and fast object detection, we are motivated to use this network for face RoI detection. We annotate 3500 video frames of facial videos and train this network  to detect the RoI as an object.  
Detecting  RoI in realistic conditions, especially when  face movement is high or some parts  of face are occluded, is very difficult.  Based on our experiments  the previous works do not work well in these conditions. The $\mathcal{R}$ is trained on videos captured in realistic conditions and  to improve the robustness.   The parameters of $\mathcal{R}$ (\ie, $\theta_{\mathcal{R}}$)  are learned to recognize the RoI of the video frames. For a video frame such as $\mathcal{F}_i$, its RoI coordinates are $\mathcal{C}_i$,  \ie, ~~ $\mathcal{R}(\mathcal{F}_i,\theta_\mathcal{R})=\mathcal{C}_i$. In summary, $\mathcal{R}$ is learned by   optimizing   the following objective function:

\begin{equation}
\mathrm{argmin}_{\theta_\mathcal{R}}{\sum_{i=1}^{K}{||(\mathcal{R}(\theta_\mathcal{R};\mathcal{F}_i)-\mathcal{C}_i)||^2}}
\label{eq:roi}
\end{equation}

where $\mathcal{C}_i$ is the coordinates of RoI in $i^{th}$ training samples and K is the number of training samples. Structure of  $\mathcal{R}$ network is  same to RFB \cite{liu2017receptive}, but it is customized and trained  for RoI task~(for more details see \cite{liu2017receptive}).

Fig. \ref{fig:roi} shows two examples of detected RoI by our learned CNNs.

\textbf{$\mathcal{S}$($\mathcal{S}$ignal Extraction):}
It is proved that the values of Red ($\mathrm{R}$), Green ($\mathrm{G}$) and Blue ($\mathrm{B}$) channels of video frames (especially $\mathrm{G}$)  are  highly correlated to  HR~\cite{de2013robust}.  Several  researches \cite{li2014remote,niu2017continuous,tulyakov2016self} are done on remote HR estimation by analyzing of color intensity of RoI pixels. 
Previous methods which are not based on deep learning,  are too slow and also are not able to run on GPU, so such methods can not be used for  applications  which need real-time process. To reduce the complexity and be able to run on GPU in real time, the  $\mathcal{S}$  analyzes  only the RoI instead of the whole frame.  Inspired by previous methods, we learn a CNN named \( \mathcal{S} \)  for understanding the
pixel color changes of RoI in respect to three different strategies:  (A) Similar to \cite{li2014remote} only the average  $\mathrm{G}$ of RoI pixels is considered, (B) The average of all three channels (\ie, $\mathrm{R}$,$\mathrm{G}$, and $\mathrm{B}$) are extracted as three signals. (C) We have divided the RoI of frames to \(h\times w \) blocks, and the average  R, G, and B of each of the blocks are computed, with which, the extracted signals are  \(3 \times h \times w \). Fig. \ref{fig:rgb_extraction} shows these three strategies for computing R, G and B signals from video frames. 


 We learn  $\mathcal{S}$ to distill the color information of a sequence of video frames as a signal. 
 Considering  strategies  explained above, we have annotated the  RoI and customized the  $\mathcal{S}$ network. As an example and for (B) case policy, We have created a set of training samples such as ${X}$ for learning $\mathcal{S}$;    $X=\{\mathcal{F}_i,(\mathrm{R}_i,\mathrm{G}_i,\mathrm{B}_i)\}_{1}^{N}$ where $F_i$ is $i^{th}$ frame and  N is total available frames.  $\mathcal{R}_i$, $\mathrm{G}_i$ and $\mathrm{B}_i$ of $\mathcal{F}_i$ are extracted with respect to traditional method such as in \cite{liu2017receptive}. $\mathcal{S}$  is leaned on X. 
 
Fig. \ref{fig:s2} shows the architecture  of $\mathcal{S}$. It includes  several convolutional  and Sub-sampling  layers and ends to a fully connected layer.    

\textbf{$\mathcal{G}$- Refinement:}
As aforementioned and can be seen in Fig. \ref{fig:os}, $\mathcal{R}$ only detects the face RoI and is robust against face movement. For robustness against noise, and low quality, the detected RoI by  $\mathcal{R}$ should be enhanced. To  this end, we learn a deep encoder-decoder network  $\mathcal{G}_1$   to refine (enhance) the  $\mathcal{R}(\mathcal{F})$. $\mathcal{G}_1$ aims to reconstruct the extracted RoI by $\mathcal{R}$, \ie,  $\mathcal{R}(\mathcal{F})$, and another CNN $\mathcal{D}_1$ which knows the distribution of high-quality RoI and supervises  $\mathcal{G}_1$ to its generated (reconstructed) sample with high quality.   The $\mathcal{G}_1+\mathcal{D}_1$ in  GAN-style  inspired by \cite{sabokrou2018adversarially} have learned. After training,  $\mathcal{G}_1$ is able to efficiently refine the RoI if it is with a low-quality or  contaminated by noise. 

With references to our experimental results,  extracted signals by $\mathcal{S}$, \ie, $\mathcal{S}(\mathcal{G}_1(\mathcal{R}(\mathcal{F})))$   probably are  contaminated with noise. \cite{li2014remote} has applied several temporal filters to smooth these signals and remove the out of range values. Similar to $\mathcal{G}_1$, we have adversarially learned an encoder-decoder  $\mathcal{G}_2$ to automatically remove the noise and out of interest range values of signals.  The $\mathcal{G}_2$ is learned in a competition of the $\mathcal{D}_2$ discriminator. $\mathcal{D}_2$ accesses the extracted signals of high-quality RoI.   

Goodfellow \etal have introduced the first version of Generative Adversarial Networks (GANs)\cite{goodfellow2014generative}. GANs tries to generate  samples that comply with  the distribution of real data, through adversarial learning of the two networks. $\mathcal{G}$ learns to map any random vector $Z$ from a latent space following a specific distribution, $p_z$, to a data sample that follows the real data distribution ($p_t$ in our case), and $D$ tries to discriminate between actual data and the fake data generated by $G$. Generator and Discriminator are learned in a two-player mini-max game, formulated as: 
\begin{equation}
\begin{aligned}
\min_{G} \max_{D} ~& \Big( \mathbb{E}_{X \sim  p_{t}}[\log(D(X))] \\ & + \mathbb{E}_{Z \sim  p_{z}}[\log(1-D(G(Z)))] \Big).
\end{aligned}
\end{equation}

Similarly, we train the $\mathcal{G}_{i}$+$\mathcal{D}_{i}$ neural networks in an adversarial procedure. Different from  earlier version of  GAN and inspired by \cite{sabokrou2018adversarially}, instead of mapping the $Z$ latent space to sample data, $\mathcal{G}$ maps a noisy version of X, denoted as  $\tilde{X}$:
\begin{equation}
\tilde{X} = \left(X \sim  p_t \right) +\left(\zeta \sim  \mathcal{N}(0, \sigma^2\mathbf{I}) \right) \longrightarrow X' \sim  p_t,
\end{equation}
where  sample X is contaminated by  noise  $\zeta$, sampled   from the Gaussian distribution with standard deviation $\sigma$, $\mathcal{N}(0, \sigma^2\mathbf{I})$.  This noise is added to the input of $\mathcal{G}$ to  make it robust to noise and improve its generalization  capability. As aforementioned, $p_{t}$ is the assumed  distribution of high-quality samples (our training samples are only high-quality ones). $\mathcal{D}$ is exposed to the high quality samples which are collected in a controlled   conditions.  Consequently it knows the  distribution of high quality samples.

$\mathcal{D}$ could decide that   $\mathcal{G}(\tilde{X})$ comes from $p_{t}$ or not. In summary, $\mathcal{G}$+$\mathcal{D}$ can be jointly learned by optimizing the following objective: 
\begin{equation}
\begin{aligned}
\min_\mathcal{G} \max_\mathcal{D} ~ & \Big( \mathbb{E}_{X \sim  p_t}[\log(\mathcal{D}(X))] \\
& + \mathbb{E}_{\tilde{X} \sim p_t+{\mathcal{N}(0, \sigma^2\mathbf{I})} }[\log(1-\mathcal{D}(\mathcal{G}(\tilde{X})))] \Big), 
\end{aligned}
\label{eq:q2}
\end{equation}

By learning $\mathcal{G}+\mathcal{D}$ based on above objective function, $\mathcal{G}$ is forced to generate samples with the distribution of $p_{t}$, Consequently, if its input  in test time is noisy or has a bad quality, it will  be automatically enhanced by  $\mathcal{G}$.

To train the model, we calculate the loss $\mathcal{L}_{\mathcal{G}+\mathcal{D}}$ as the loss function of the joint network $\mathcal{G}$+$\mathcal{D}$. Besides, we need $\mathcal{R}$'s output to be close to the original input image. As a result, an extra loss is imposed on the output of $\mathcal{R}$:
\begin{equation}
\mathcal{L}_\mathcal{G}=\| X-X'\|^{2}.
\end{equation}

Therefore, the model is optimized to minimize the loss function:
\begin{equation}
\mathcal{L}=\mathcal{L}_{\mathcal{G}+\mathcal{D}}+ \lambda \mathcal{L}_\mathcal{G},
\label{eq:sum_loss}
\end{equation}

Where $\mathcal{L}_{\mathcal{G}+\mathcal{D}}$  is calculated based on Equ. \ref{eq:q2}. After adversarial training of $\mathcal{G}_i$ and $\mathcal{D}_i$, the $\mathcal{G}_i$ is able to improve the quality of its input.  We learn two encoder-decoder networks for refining the intermediate results of our targeted task. $\mathcal{G}_1+\mathcal{D}_1$ and $\mathcal{G}_2+\mathcal{D}_2$ are learned on very high-quality RoI of face and extracted information (signals) from these RoIs, respectively. Therefore, in testing, $\mathcal{G}_1$ and $\mathcal{G}_2$ are two efficient yet simple neural network for refining the output of $\mathcal{R}$ and $\mathcal{S}$, respectively. 

\begin{figure}[t]
    \centering
   \includegraphics[width=0.6\linewidth]{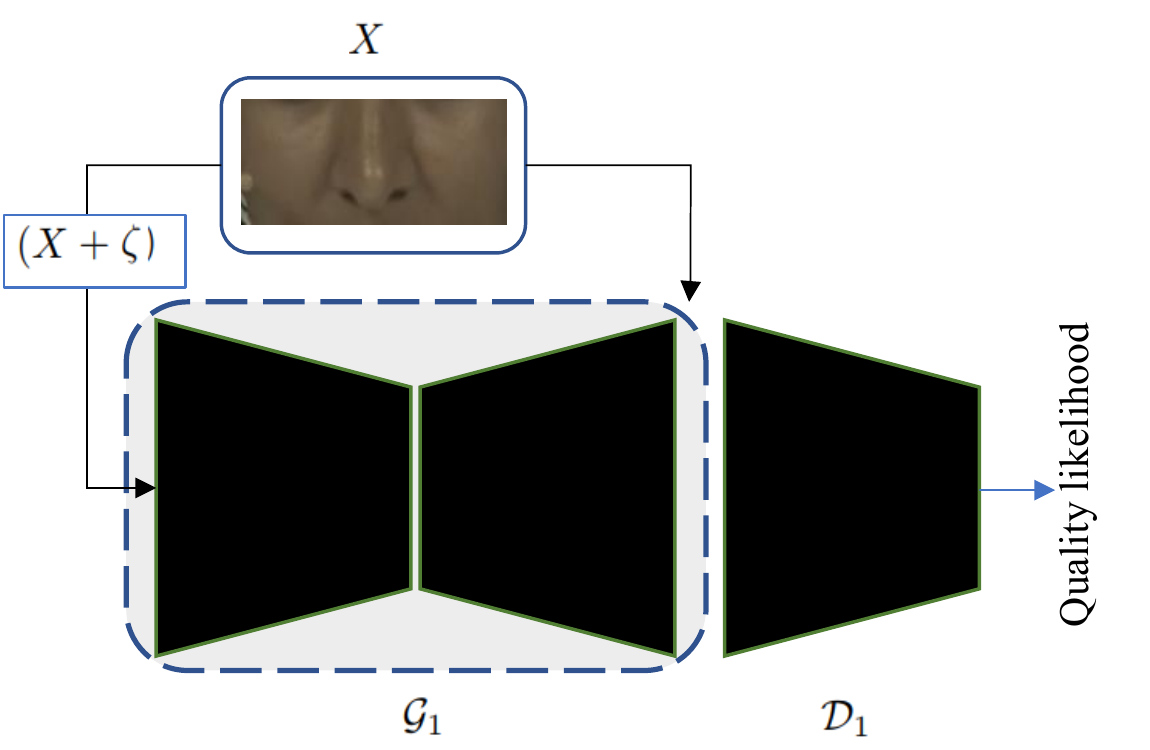}
    \caption{A sketch of $\mathcal{G}_1+\mathcal{D}_1$ networks.  After training,  $\mathcal{G}_1$ is a RoI refiner.}
    \label{fig:g1d1}
\end{figure}

Fig.~\ref{fig:g1d1} shows a sketch of $\mathcal{G}_1 + \mathcal{D}_1$ network. Due to limited space, detailed architecture  and learning parameters of the  $\mathcal{G}_{k}+\mathcal{D}_{k}$ (K={1,2}) are supplemented to this paper.   

\subsection{BE: Regression} The \textbf{FE}  is devised to efficiently distill the facial videos as an informative representation which is highly correlated to HR. Therefore, learning to estimate  HR relying  on  new representation (\ie, $\mathcal{G}_2(\mathcal{S}(\mathcal{G}_1(\mathcal{R}(\mathcal{F}))))$) is not as difficult as understanding from raw videos, and could be learned by a light deep auto-encoder on limited samples.  In the testing  stage, for capturing the subtle  color changes, the detected RoIs by $\mathcal{R}$ from frames of a video with length T,  pass   to $\mathcal{S}$ one by one,  then $\mathcal{S}$ map the RoI to informative signals    $Z\in \mathbb{R}^{3\times T}=\{\mathrm{R}_i$, $\mathrm{G}_i$, $\mathrm{B}_i\}_{i=1}^{T}$. As mentioned, we enhance the quality of these signals by  $\mathcal{G}_2$  refiner network as the final output of $\textbf{FE}$. 

Mapping the Z$\in\mathbb{R}^{3 \times T}$ signals to a continues value such as HR can be defined as a regression problem.  To solve this regression problem,  simple yet efficient DNNs are learned on  labeled training samples.  If we have  K annotated video clips for training, then our training set is $\mathbb{M}$=\{\textbf{FE}($\mathcal{V}_i$)=$Z_i$, Y$i$)$^{i=1:K}$\} where Y$_i$ is the average HR of i$^{th}$ video clip. 

In summary, a light DNN is learned on $\mathbb{M}$ to maps the $Z_i$ to its equivalent HR, \ie, Y$_i$.  This estimator network ($\mathbb{E}$) is trained by optimizing the below objective function. (see Equ.\ref{eq:MSE}) 

\begin{equation}
    \frac{1}{K}\sum_{i=1}^{K}{||Y_i-\tilde{Y}_i||^2}
    \label{eq:MSE}
\end{equation}

Where $\tilde{Y}_i=\mathbb{E}(\mathcal{V}_i)$  and Y$_i$ is the label  (\ie, HR) of i$^{th}$ training video. 

\begin{figure}[t]
    \centering
   \includegraphics[width=.4\linewidth]{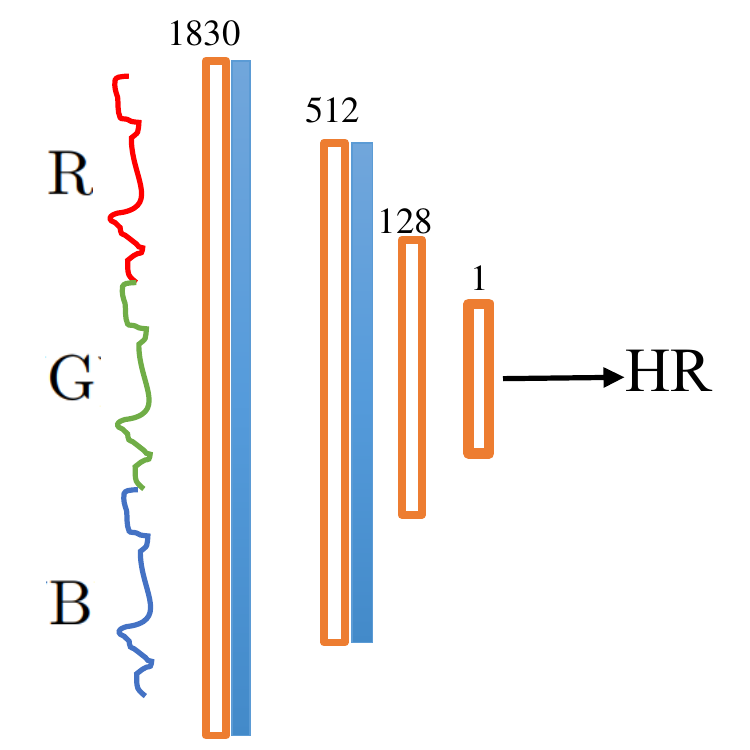}
    \caption{The architecture of $\mathbb{E}$ for HR estimation. All layers are fully-connected, and batch normalization is shown by blue}
    \label{fig:regression}
\end{figure}

The $\mathbb{E}$ network  includes three fully-connected layers $\{fc_1, fc_2, fc_3\}$ with size of 512, 128 and 1.  A  batch normalization layer~\cite{ioffe2015batch} is embedded after two first layers. A sketch of this network is shown in Fig \ref{fig:regression}.

\section{Experimental Results}
\label{sec:er}
In this section, We evaluate the performance of the proposed method, and also the results of our method are compared with state-of-the-art methods for remote HR estimation from videos.   Most of the state-of-the-art methods on HR estimation have only focused to achieve an accurate result. One important aspect which is neglected by them is the run-time. We also carry out several experiments to analyze the run-time of our method. Results confirm that Deep-HR method can work in real time  and accurately compute the HR rate from videos which are captured in realistic conditions.

\subsection{Experimental setting}
Our method is implemented  using PyTorch~\cite{paszke2017automatic}  framework on a GeForce GTX 1080 GPU.  The $\mathcal{R}$, $\mathcal{S}$ and $\mathbb{E}$ are three important networks which are used for HR estimation. The former two networks form \textbf{FE} and latter one as \textbf{BE}.  The $\mathcal{R}$ are trained on 3500  face images which are randomly picked  from datasets containing  face images( Here we only use from MAHNOB and HR-D datasets).  We have carefully labeled the RoI of each  face image. The size of RoIs is  $224\times 74$ pixels. The learning rate, weigh decay and momentum of the $\mathcal{R}$ are 0.004, 0.00005 and 0.9, respectively.  We also extract the R, G and B signals from the RoI of these 3500 images as training samples  $\mathcal{S}$ network. The learning rate of $\mathcal{S}$  is 0.01 and its momentum equals to 0.9. 
The \textbf{BE}, \ie, $\mathbb{E}$ network is learned by learning rate of 0.01. 
In Equ. \ref{eq:sum_loss}, the $\lambda$ hyper-parameter for learning of $\mathcal{G}$ network equals to 0.2. 
More details about the architectures, learning parameters related to the refiner network are explained in supplementary material of this paper.

\subsection{Evaluation measures}
Researchers have exploited different statistical measures for evaluating the efficiency of their proposed methods. To be fair and have a comprehensive comparison with state-of-the-art methods,  five standard    measures are utilized: 

 (1)$M_{\mathcal{D}}$, (2)SD$_{\mathcal{D}}$, (3)RMSE$_{\mathcal{D}}$  , (4) Me$_{\mathcal{D}}$, and (5) ${r}$.
 
 $\mathcal{D}_i=Y_i-\tilde{Y}_i$, where Y$_i$ and $\tilde{Y}_i$ are the estimated HR and ground truth of $i^{th}$ test video clip, respectively.   $M_{\mathcal{D}}$ is the average of errors and equals  $\frac{1}{K}\sum_{i=1}^{k}D_i$ where K is the number of test samples.  $SD_{\mathcal{D}}$ is the standard deviation of errors.   The  RMSE$_{\mathcal{D}}=\sqrt{\frac{1}{K}\sum_{i=1}^{i=K}{||Y_i-\tilde{Y}_i||^2}}$ is an efficient way to evaluate the errors of a method.  The fourth  measure, \ie, Me$_{\mathcal{D}}$ is the  mean of error-rate  percentage: Me$_{\mathcal{D}}=\frac{1}{K}\sum_{i=1}^{K}\frac{\mathcal{D}_i}{\tilde{Y}_i}$, as mentioned in previous the $\tilde{Y}_i$ is ground truth of $i^{th}$ video. The last measure  $r$,  is the linear correlation between estimated signal $(\{{Y}_i\}_{i=1}^{N})$ and  ground truth signal $(\{{\tilde{Y}}_i\}_{i=1}^{N})$ assessed using Pearson’s correlation coefficients $r$ and its  $p$
value. Pearson’s  $r$ varies between -1 and 1, where  $r$ = 1  indicates total positive correlation and r = -1 indicates total negative correlation. The p value is the probability of the statistical significance of the test  if the calculated r were in
fact zero (null hypothesis). Usually the result is accepted as statistically significant when p $\leq$ 0.01. 

\subsection{Datasets}
In previous researches for remote HR estimation, results on several datasets are reported. Almost most of these datasets are not public. In another side, as mentioned previously proposed method (except \cite{niu2018synrhythm}) for remote estimation of  HR do  not rely on deep learning, and they do not exploit a set of labeled data for training. This brings some difficulties for us to evaluate and compare our method with state-of-the-art methods. To this end, we evaluate our method on a public and popular dataset, MAHNOB~\cite{soleymani2012multimodal} for this task and we also introduce a new dataset named HR-D for HR estimation.

\textbf{MAHNOB:} This data-set is collected from 27 subjects (15 females and 12 males). 20 high-resolution clips are captured from every subject.  
Each subject participated in two different experiments: emotion elicitation and  implicit tagging. 
We follow the  setting in \cite{li2014remote} for doing our experiment on this dataset. We have used interval size of 30 seconds (306$^{th}$ to 2135$^{th}$ frames) from 527 videos. The second channel  (EXG2) of the  corresponding ECG wave-forms  is exploited as ground truth for computing the heart rate.

\begin{figure}[t]
    \centering
   \includegraphics[width=0.5\linewidth]{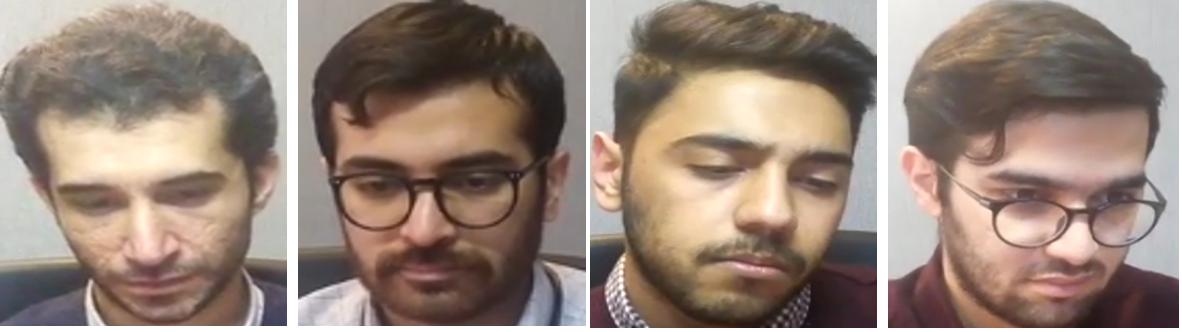}
   \caption{Four examples of our dataset~(HR-D)}
    \label{fig:od}
\end{figure}

\textbf{HR-D:}  To create this dataset, we have recorded  75 videos with an average of   82 seconds duration and 22 frames per second. The resolution of the video frames is 800$\times$ 480. Subjects ranging from 22 to 33 years old participate for collecting  this dataset. To make the video  realistic and challenging, the videos are recorded from different angles, distance and  poses. The Videos were captured while the participants were watching the different genre of movies. Figure \ref{fig:od} shows several frames of created dataset. As can be seen, some videos were recorded from the participant with mustaches or bearded  that makes the HR estimation task from videos very challenging.        
 

\subsection{Results}
\textbf{Results on MAHNOB:}
As mentioned, since the annotated samples of MAHNOB dataset are very limited,  dividing these data to training and testing samples are not reasonable. Instead, K-fold cross-validation with K=3 is exploited for evaluation. It is worth mentioning that we only use these samples to train the $\mathbb{E}$ network. The $\mathbb{E}$ does not train on the raw sample since  the \textbf{FE} maps the raw frames to an informative representation and  $\mathbb{E}$ is trained on it. 

Table \ref{tab:res_mah} shows the performance of our method with comparison to state-of-the-art methods.  As we explained in Section \ref{sec:fe}, three different policies (A, B and C) are considered for mapping the facial video frames to informative representation as the output of $\textbf{FE}$.  The results in Table \ref{tab:res_mah} show that  the A strategy, \ie, computing the average of green color, obtains the best performance.  

Note that \cite{niu2018synrhythm} and \cite{vspetlikvisual}  are also based on deep learning. As can be seen, our method get substantially  better results than these two methods  by a considerable margin. The \cite{niu2018synrhythm} only uses a simple deep network, similar to  which we use as \textbf{BE}, while  \cite{vspetlikvisual}    captures the subtle color changes from whole face (not from RoI) without any refining. Notably, we extract the  subtle color changes in two steps, and enhance our output in each of steps. Based on our experiment, very deep network fails to interpret the subtle color changes. Consequently,  amplification  such as $\mathcal{G}$ is necessary for  enhancing the  intermediate out of these procedures.  

The outcome of our deep network for estimation of HR is more accurate than other compared  methods in Table \ref{tab:res_mah}. It  demonstrates that our method can accurately estimate the HR  with just 3.41 RMSE, and 0.0273 Me$_\mathcal{D}$.  

\begin{table}[t]
\centering  
    \caption{Comparison of Deep-HR method with state-of-the-art methods  for HR estimation on MAHNOB dataset. Numbers in bold show the best results.}
\begin{tabular}{lcccl}
\hline
 Methods & M$_\mathcal{D}$ /SS$_{\mathcal{D}}$ & RMSE$_{\mathcal{D}}$ & Me$_{\mathcal{D}}$&${r}$   \\\hline \hline
 Poh~\etal~\cite{poh2010non} &-8.95  /24.3 & 25.9 & 25.0\%& 0,08\\ 
Poh~\etal\cite{poh2011advancements}& 2.04/13.5 &13.6 &13.2\%&   0.36\\
\scriptsize{Balakrishna~\etal}\cite{balakrishnan2013detecting} & -14.4/15.2& 21.0 &20.7\% & 0.11 \\
Li~\etal\cite{li2014remote} & -3.30/6.88& 7.62& 6.87\%&0.81\\
\scriptsize{De Haan~\etal} \cite{de2013robust}  &-2.89/13.67 &10.7 &12.9\%&0.82\\
\scriptsize{Tulyakov~\etal} \cite{tulyakov2016self} &3.19/5.81 &6.23 &5.93\%&0.83\\
Hus~\etal \cite{hsu2014learning} & \textbf{-0.20}/ 11.32 &11.31& 12.8\%&--\\  \hline \hline
Niu~\etal\cite{niu2018synrhythm}  & 0.30/4.48& 4.49& 4.37\% &--\\
\scriptsize{Spetlik} \etal\cite{vspetlikvisual} & 7.25/-- &9.24&--&0.51 \\  \hline \hline

Ours(C) & -2.16/ 4.13 &4.07& -0.0288 & 87\%\\
Ours(B) & 3.11/ 3.61 &3.56&0.042& 91\%\\
Ours(A) & 2.08/\textbf{3.47} &\textbf{3.41}&\textbf{0.0273}& \textbf{92}\%\\
    \end{tabular}
    \label{tab:res_mah}
\end{table}
\textbf{Results on HR-D:}
To show the efficiency of the proposed method, we evaluate it on HR-D dataset. Due to some issues such as personal privacy and high  cost of   video labeling  for HR estimation task, collecting numerous labeled training samples is very challenging  and, consequently the training data  for this task is very limited.  In such a case, \ie, scarcity of labeled samples,  we use K-fold cross-validation  (K=3) to assess the performance of our method (Similar to \cite{niu2017continuous}). Table \ref{tab:res_our} shows the performance of our method.  Better results are achieved  by a remarkable margin with respect to different measures. It appears that the previous methods could not efficiently estimate the HR from face videos of people with mustaches or bearded. It is because that they are very sensitive to the  pixel colors of RoI, and such things like to mustaches or bearded collapse the extracted signal from RoI.  Our deep network for HR estimation is a trainable approach and can simply adapt for different variations of the face. Furthermore, if there is  small distortion on RoI or extracted signals, the $\mathcal{G}_1$ and $\mathcal{G}_2$ are able to refine and improve the quality of data.   
\begin{table}[t]
\centering  
    \caption{Comparison of Deep-HR method with  recently published  state-of-the-art methods  for HR estimation on HR-D. Numbers in bold show the best results.}
\begin{tabular}{lcccl}
\hline
 Methods & M$_\mathcal{D}$/SD$_{\mathcal{D}}$ & RMSE$_{\mathcal{D}}$ & Me$_{\mathcal{D}}$&${r}$   \\\hline \hline
 Hus \etal & -4.17/25.02 &24.71&-0.054& 17.19\%\\  
 Tulyakov \etal  &8.32/14.68 &15.13 &0.1156 &12.94\%\\
 Niu \etal& -2.5/17.51 &14.02 & -0.0301& 24\%\\
Ours &\textbf{3.02}  / \textbf{7.14}  &\textbf{6.58} &\textbf{0.0343}& \textbf{81}\% \\ 
    \end{tabular}
    \label{tab:res_our}
\end{table}

\begin{figure}[pt]
    \centering
   \includegraphics[width=0.7\linewidth, height=0.25\linewidth]{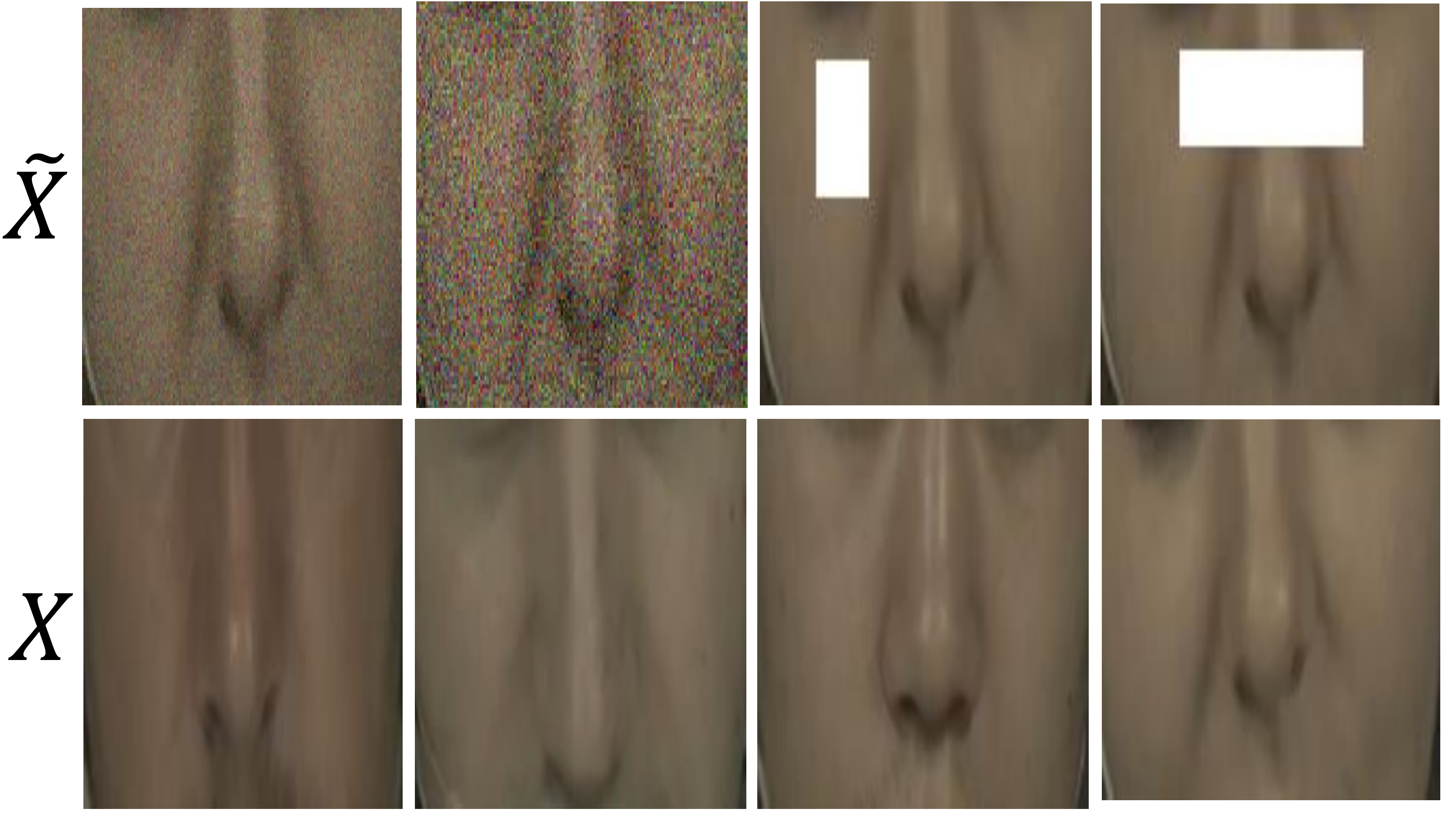}
    \caption{Some examples of refined samples by $\mathcal{G}_1$. $\tilde{X}$: low-quality samples, $X=\mathcal{G}_1(\tilde{X})$: Refined samples by $\mathcal{G}_1$}
    \label{fig:g1g2}
\end{figure}

\textbf{Refining Results:}
The success of our proposed method is largely due to quality improvement of  the output of $\mathcal{R}$ and $\mathcal{S}$ deep networks by $\mathcal{G}_1$ and $\mathcal{G}_2$, respectively.  Fig. \ref{fig:g1g2} is a showcase of the output of $\mathcal{G}_1$ refiner networks. As can be seen, the refiner network  efficiently improve the quality of the  proposed method. For first two  images  in the top row it works as an de-noising network and for to next ones, as a in-painting network  (first two in bottom row).

\textbf{Run-time:}
In most  HR estimation applications especially in health-care, real-time HR estimation  is very vital. Most of the previous works is too complex to run in a real time. They did not discuss the computational  complexity (or run time) of their methods. Consequently,   direct comparison of our method with other state-of-the-art methods is impossible.   Our results  on  a GeForce GTX 1080 GPU shows that our method can run as fast as 100 frames per second. It  confirms that our method is able to estimate HR very fast.

%


\section{conclusion}
\label{sec:con}
In this paper, we proposed a learning framework for fast and accurate HR estimation, which is able to learn from limited training samples. Our proposed method is composed of two pivotal components: $\textbf{FE}$ and $\textbf{BE}$. $\textbf{FE}$ improves the interpretability of subtle color changes of facial videos, while  \textbf{BE}   estimates HR from output of \textbf{FE}. Two  refiner networks, \ie, $\mathcal{G}_1$ and $\mathcal{G}_2$ are care about the quality of    intermediate and final output of $\textbf{FE}$. These two refiners are adversarially learned to understand the distribution of high quality RoIs and extracted color signals from RoIs.  We evaluate our method on two datasets.  The results suggest that our method is superior to or at least comparable with the state-of-the-art on those datasets while being able to run in real time.

\clearpage
%
%
\bibliographystyle{splncs04}
\bibliography{egbib}
\end{document}